\documentclass{article} 
\usepackage[utf8x]{inputenc}
\usepackage{nips15submit_e,times}
\usepackage{hyperref}
\usepackage{url}
\usepackage{amsmath}
\usepackage{amsfonts}

\usepackage{array}
\usepackage{graphicx}
\usepackage{algorithm}
\usepackage{algorithmic}
\usepackage{mathtools}

\DeclarePairedDelimiter\floor{\lfloor}{\rfloor}
\nipsfinalcopy

\newcommand{\Expect}{\mathbb{E}}
\newcommand{\Var}{\mathbb{V}}

\newcommand{\acor}{\rho} 

\newcommand{\targfunc}{h} 

\newcommand{\targd}{f} 
\newcommand{\propd}{q} 

\newcommand{\smp}{{X}} 

\title{Gradient Importance Sampling}

\author{
Ingmar Schuster\\
Natural Language Processing Group \\
Department of Computer Science\\
University of Leipzig\\
\texttt{schuster@informatik.uni-leipzig.de} \\
}

\begin{document}

\maketitle

\begin{abstract}
Adaptive Monte Carlo schemes developed over the last years usually seek to ensure ergodicity of the sampling process in line with MCMC tradition. This poses  constraints on what is possible in terms of adaptation. In the general case ergodicity can only be guaranteed if adaptation is diminished at a certain rate. Importance Sampling approaches offer a way to circumvent this limitation and design sampling algorithms that keep adapting.
Here I present a gradient informed variant of SMC (and its special case Population Monte Carlo) for static problems.
\end{abstract}

\section{Introduction}

Monte Carlo methods have been developed into one of the mainstream inference methods of Bayesian Statistics and Machine Learning over the last thirty years \cite{Robert2004}. They can be used to approximate expextations with respect to posterior distributions of a Bayesian Model given data. The most widely used Monte Carlo Method for this purpose today is Markov Chain Monte Carlo (MCMC). In this approach, a Markov Chain is constructed which is ergodic with respect to the given posterior distribution.
In parallel, a scheme for sampling from a sequence of target distributions, called Sequential Monte Carlo (SMC), has been developed \cite{Doucet2001a}. SMC has traditionally been used for inference in time series models and for online tracking applications. However, there has been a considerable amount of research on using SMC for inference in static models as well \cite{Chopin2002,DelMoral2006,Schafer2013}. In this paper, I will develop a powerful variant of SMC for static models making use of gradient information, dubbed Gradient Importance Sampling.

The paper will proceed as follows. In subsection \ref{sec:lmc} I will give a short overview of the a simple well-known MCMC algorithm making use of gradient information, the Metropolis Adjusted Langevin Truncated Algorithm. In subsection \ref{sec:ismc} an exposition to Importance Sampling and SMC is given. Gradient IS and its variants  are introduced in \ref{sec:GRIS}. Related work, especially in adaptive MCMC and Importance Sampling, is discussed in Section \ref{sec:relwork}. Gradient IS is evaluated and compared against previous algorithms in Section \ref{sec:eval}. The last section concludes.

\subsection{Langevin Monte Carlo}
\label{sec:lmc}
Metropolis Adjusted Langevin Truncated Algorithm (MALTA) \cite{Roberts1996} and Hamiltonian Monte Carlo (HMC) \cite{Neal2011} are two well known sampling algorithms that make use of gradient information. In the following, we will denote the target density as $f$. Often times, this will we the posterior of a Bayesian model which can be evaluated only proportionally by multiplying the prior and likelihood at a given a point. \\
 HMC is probably better known in the Machine Learning community, but it is notoriously complex and its description is beyond the scope of this pape. For a thorough introduction see e.g. \cite{Neal2011,Bishop2007}. The special case of HMC however, MALTA, is closely related to the algorithm proposed in this paper and a concise introduction will be given. MALTA is a variant of the Metropolis-Hastings MCMC algorithm where, given the current state of the Markov Chain $\smp'$, a proposal for a new state $\smp$ is sampled from the multivariate normal density
$$q(\cdot | \smp') = N(\cdot | \smp'  + D(\nabla~\textrm{log}~f(\smp' )), C)$$
where $D$ is a drift function. For consistency reasons, the MALTA variant 
used in the evaluation section will use $D(\nabla~\textrm{log}~f(\smp' )) = \delta\nabla~\textrm{log}~f(\smp')$ for $0 \leq \delta \leq$ 1. The covariance matrix $C$ is fixed by the user prior to running the algorithm.
The proposed new state $\smp$ is then accepted with the usual Metropolis-Hasting acceptance probability $$min\left(1,\frac{f(\smp')q(\smp | \smp')}{f(\smp)q(\smp' | \smp)}\right)$$ and recorded as a sample.
 If the proposed state is rejected, the chain remains at state $\smp'$ (which is recorded as a sample again). The Markov Chain is ergodic with respect to $f$, i.e. the samples produced are approximately from $f$, which is guaranteed by using the Metropolis-Hastings correction. The samples can be used to estimate the expectation $H$ of some function of interest $h$ with respect to the target density $f$ using the law of large numbers as
 $$H = \int h(x) f(x) \textrm{d}x \approx 1/N \sum_{i=1}^Nh(\smp_i)$$
where $\smp_i$ ranges over the samples and $N$ is the number of samples.

\subsection{Importance Sampling and SMC}
\label{sec:ismc}
Importance Sampling takes a different approach. Instead of trying to sample approximately from $f$, it samples from some proposal density $q$ instead. Rather than correcting for the change of distribution using Metropolis-Hastings, the Importance Sampling estimator simply weighs each sample $\smp$ by the so called importance weight $w(\smp) = f(\smp)/q(\smp)$. In the case where $f$ is not normalized, which is the usual case when estimating a Bayesian posterior, the self-normalized Importance Sampling estimator for $H$ given by

 $$H = \int h(x) w(x) q(x) \textrm{d}x \approx \frac{1}{\sum_{i=1}^N w(\smp_i)} \sum_{i=1}^N w(\smp_i) h(\smp_i)$$
 
Sequential Monte Carlo (SMC) \cite{Doucet2001a}  builds on Importance Sampling and was originally devised to sample from a sequence of target distributions. For ease of exposition, I will first consider the case where the same target distribution is used at each iteration, a special case known as Population Monte Carlo (PMC). From this, an extension to  sequence of targets is straight forward and given in Section \ref{sec:seqin} for the case of static models (i.e. not time series).
In Population Monte Carlo, we first gather a set of $p$ samples (also called \emph{particles} in SMC) $\smp_1,\dots,\smp_p$ from proposal densities $q_1,\dots,q_p$ which are assigned weights $w(\smp_i)=f(\smp_i)/q_i(\smp_i)$. Instead of using these weighted samples directly with the Importance Sampling estimator to evaluate the integral of interest, we resample  $\smp_1,\dots,\smp_p$ with replacement according to their respective weights, adding the resulting set to a set of unweighted samples $S$. This is called Importance Resampling and produces a sample set that is approximately coming from the posterior \cite{Rubin1987}. Several methods exist for this step, the easiest being multinomial resampling. See \cite{Douc2005} for a review including some theoretical results. 
Previous samples can now be used to construct proposal distributions for the next iteration.
In the simplest case this could be centering a proposal distribution on a previous sample. The procedure is iterated until $S$ is deemed large enough. The integral of interest can now simply be computed by
$$H = \int h(x) f(x) \textrm{d}x \approx 1/|S| \sum_{\smp \in S}^Nh(\smp)$$
Moreover, the marginal likelihood $Z$ of the data (also called evidence of the model or normalizing constant of $f$) can be approximated by the formula
$$Z \approx 1/N_w \sum_{i=1}^{N_w} w_i$$
where $w_i$ are the weights that have been gathered from the stage before resampling and $N_w$ is the total number of weights.

A major argument for Gradient IS is the ability to approximate the marginal likelihood \emph{and} the target distribution as good as or better than previous gradient-informed and/or adaptive sampling algorithms while being extremely simple to implement. For example, this opens the possibility to routinely compute Bayes factors (and thus do Bayesian Model selection) as a by-product of very  efficient posterior sampling instead of using special inference techniques geared towards only computing $Z$.

\section{Gradient IS}
\label{sec:GRIS}
Gradient IS (GRIS) is a variant of Sequential Monte Carlo \cite{Doucet2001a}, or, when targeting the same density at each iteration, of its special case Population Monte Carlo \cite{Cappe2004}. GRIS accomplishes adaptivity by  fitting a covariance matrix $C_t$ to samples from the target density that have been gathered at time $t$. The proposal distribution for a new sample given an old sample $\smp'$   is then given by
\begin{equation*}
q_{t}(\cdot | \smp' ) = N(\cdot | \smp'  + D(t, \nabla~\textrm{log}~f(\smp' )), C_t)
\end{equation*}
 where $D$ is a drift function. We used $$D(t,  \nabla~\textrm{log}~f(\smp' )) = (\delta /t^{1.5})  \nabla~\textrm{log}~f(\smp' )$$ thus introducing a parameter $\delta \geq 0$ (and usually $\delta \leq 1$) which ought to be tuned for each target density. 
To fit $C_t$ we use the parametrization
\begin{equation*}
C_t = \begin{cases}
C_0 & t \leq t_0 \\
s_d (cov(X_0,\dots, X_{t-1}) + \epsilon I)& t > t_0
\end{cases}
\end{equation*}
For an initial $C_0$ and some $t_0$ and where $s_d>0$ and $\epsilon>0$ are tunable parameters. To update $C_t$ with a new sample $X_t$, we can a recursion formula 
\begin{equation*}
C_{t+1} = \frac{t-1}{t} C_t + \frac{s_d}{t} \left (t\bar{X}_{t-1}\bar{X}^T_{t-1} - (t+1) \bar{X}_{t}\bar{X}^T_{t} + {X}_{t}{X}^T_{t}  + \epsilon I\right )
\end{equation*}
Where $\bar{X}_t$ is the running average at iteration $t$ (with an obvious recursion formula) \cite{Haario2001}.
Directly updating either the Cholesky or Eigendecomposition of $C_t$ results in significant computational savings, especially when the updates are done frequently. Both decompositions can be used to draw samples from and evaluate the density of the respective multivariate normal, both of which are prerequisites of the GRIS algorithm. A complementary method to tradeoff speed and adaptivity is to collect more than one new sample before updating. GRIS then proceeds as given in  Algorithm~\ref{algo:lis}.

\begin{algorithm}[tb]

\caption{Gradient IS algorithm}
\begin{algorithmic}
\label{algo:lis}
   \STATE {\bfseries Input:} unnormalized density $f$, gradient $\nabla~\textrm{log}~f$, population size $p$, $p$ intial samples $S$, sample size $m$
    \STATE {\bfseries Output:} list $S$ of $m$ samples
   \WHILE{$len(S) < m + p$}
   	\STATE Initialize $P = List()$
	\STATE Initialize $W = List()$
	\FOR{$i=1$ {\bfseries to} $p$}
		\STATE (a) sample $\smp' $ uniformly from last $p$ samples in $S$
		\STATE (b) generate $\smp \sim q_{t}(\cdot | \smp' )$, append it to $P$
		\STATE ~~~append weight $f(\smp)/q_{t}(\smp)$ to $W$
   	\ENDFOR
	\STATE resample $p$ values from $P$ with repl. according 
	\STATE ~~~to weights and append samples to $S$
	\STATE compute $C_{t+1}$

   \ENDWHILE

   \STATE remove first $p$ samples from $S$
\end{algorithmic}

\end{algorithm}

\subsection{Gradient IS with a sequence of target distributions}
\label{sec:seqin}
Instead of using the correct target distribution $f$ at every IS iteration like in PMC, it is also possible to construct a sequence of intermediate target distributions $(g_t)_{t=1}^T$ where $g_T = f$. In the existing SMC literature addressing static targets, only the samples acquired when targeting the actual distribution of interest are then used for estimating the integral $H$ \cite{DelMoral2006,Schafer2013}. This approach is very robust when compared to MCMC algorithms, as it explores the probability space better \cite{Schafer2013,Chopin2002}.\\
One possibility for constructing a sequence of distributions is the geometric bridge defined by $g_t \propto g_0^{1-\rho_t} f^{\rho_t}$  for some initial distribution $g_0$ and where $(\rho_t)_{t=1}^T$ is an increasing sequence satisfying $\rho_T = 1$. Another is to use a mixture $g_t \propto ({1-\rho_t})g_0+ {\rho_t}f$. When $f$ is a Bayesian posterior, one can also add more data with increasing $t$, e.g. by defining the intermediate distributions as $g_t(\smp) = f(\smp|d_1,\dots,d_{\floor{\rho_t D}})$ where $D$ is the number of data points, resulting in an online inference algorithm \cite{Chopin2002}.

However, when using a distribution sequence that computes the posterior density $f$ using the full dataset (such as the geometric bridge or the mixture sequence), one can reuse the intermediate samples when targeting $g_t$ for posterior estimation using a simple trick. As the value of $f$ is computed anyway for the harmonic bridge and the mixture sequence, we can just use the weight $f(\smp)/q_t(\smp)$ for posterior estimation while employing $g_t(\smp)/q_t(\smp)$ to inform future proposal distributions. This way the evaluation of $f$ (which is typically costly) is put to good use for improving the posterior estimate. To the best of my knowledge, this recycling of samples in SMC for static targets has not been reported in the literature.

\section{Related work}
\label{sec:relwork}

\subsection{Adaptive Monte Carlo using ergodic stochastic processes}
Adaptive MCMC algorithms use samples acquired so far in order to tune parameters of the sampling algorithm to adapt it to the  target.  The first algorithm of this class, the Adaptive Metropolis (AM) Algorithm \cite{Haario2001} fits a Gaussian approximation with mean $\bar{\smp}_{t}$ and covariance matrix $C_t$ to the samples acquired up until iteration $t$,  in exactly  the same fashion as GRIS. The proposal distribution then is the same as that of GRIS, except for the fact that no gradient information is used, i.e. the drift function is $D(\cdot,\cdot) = 0$. A proposal then is accepted or rejected using the Metropolis-Hastings correction, which is another difference from GRIS, which uses importance weighting instead. The AM algorithm is the first adaptive MCMC algorithm and is provably ergodic wrt the target distribution $f$.

A similar algorithm making use of gradient information is the adaptive Metropolis Adjusted Langevin Algorithm with truncated drift (adapt. MALTA) \cite{Atchade2006}. Adaptive T-MALA uses a proposal distribution given by $N(\cdot | \smp_{t-1} + \frac{C_t}{2} D(\smp_{t-1}), C_t)$, where $D(\cdot)$ is a drift function. The covariance $C_t$ is constrained to lie in the convex compact cone defined by the projection $p_2(C_t') = C_t'$ if $|C_t'| \leq A_1$ and $p_2(C_t') = \frac{A_1}{|C_t'|}C_t'$ else for some parameter $A_1$ and where $|\cdot|$ is the Frobenius norm. Similar constraints and accompanying projections are given for adapting the $s_d$ parameter and fitting the target sample mean $\bar{\smp}_{t}$ (for details see \cite{Atchade2006}).

Informally, adapting the Markov Kernels used in MCMC will result in an algorithm that is ergodic with respect to target $f$ in the general case as long as 
\begin{enumerate}
	\item all Markov kernels used are ergodic wrt target $f$ (have $f$ as their stationary distribution)
	\item adaptation diminishes, e.g. adaptation probability $p_t$ for iteration $t$ satisfies $p_t~\rightarrow~0$ as $t~\rightarrow~\infty$.
\end{enumerate}
(Theorem 1 in \cite{Roberts2007}). We can still have $\sum^N_{t=1}p_t {\rightarrow} \infty$ as $N\rightarrow\infty$, i.e. infinite overall adaptation (as in \cite{Sejdinovic2013}). It is important to note that the diminishing adaptation is sufficient but not necessary. Other schemes might still be ergodic, one example being the AM algorithm, where adaptation is not diminished.

\subsection{Adaptive Importance Sampling Schemes}

Adaptive Importance Sampling schemes in related to SMC have mostly been trying to fit a closed-form approximation to the posterior for usage as a proposal distribution in the next iteration. In particular, previous work used optimality criteria such as minimization of KL-divergence between proposal distribution and posterior to asses the quality of the proposal distribution. After collecting new samples, D-Kernel approximations \cite{Douc2007} or Mixture distributions (such as Mixtures of Gaussians or Student-T mixtures) \cite{Cappe2008} are updated to fit the posterior better.
In another line of research, better ways of using the produced samples have been sought. A particularly interesting approach is Adaptive Multiple Importance Sampling \cite{Cornuet2012,Marin2012}. Here, samples from previous proposal distributions are reweighted after the sampling is finished in order to reflect the overall proposal distribution used across iterations. This is achieved by simply assuming that each weighted sample is produced by a mixture over the proposal distributions used, where the mixture weight is proportional to the number of samples actually drawn from a particular proposal. This reweighting has been shown to be consistent and considerably improve consecutive estimates \cite{Marin2012}.

\section{Evaluation}
\label{sec:eval}
For an experimental evaluation of the adaptive Gradient IS algorithm, I targeted three synthetic distributions (i.e. not posteriors based on some dataset) and one posterior of a Bayesian Logistic regression model. The synthetic distributions have the advantage that the expected value $H$ of the target is known in closed form, for the Logistic Regression dataset ground truth was estimated in a separate Monte Carlo experiment. I started the simulation at $H$ to avoid having to handle burn-in for MCMC and ran $20$ Monte Carlo simulations with $3000$ target or gradient evaluations each. If the target and its gradient are evaluated at the same point, this is counted as a single evaluation. This is justified by the fact that most of the time either likelihood or gradient calculation dominate complexity and computing the other value uses only little additional time (since factors common to both likelihood and gradient can be cached).  I compared to the AM algorithm, adaptive T-MALA, and Hamiltonian Monte Carlo \cite{Neal2011} MCMC algorithms, where I tuned the parameters to give optimal acceptance rates as reported in the literature. GRIS was tuned for low estimates of Monte Carlo variance, an information that is available in typical black-box situations (i.e. where nothing is known about the target $f$ except the information produced by the sampling algorithm). Is used the variant of GRIS that did not use a sequence of target distributions as it gave good results and was easier to implement.

\subsection{Maximum squared errors and Effective Sample Size}
Estimating  Effective Sample Size (ESS) is easy in traditional Importance Sampling and does not require establishing ground truth. Given importance weights of $N$ collected samples $(w_i)_{i=1}^N$ and their normalized version $\widehat{w}_i = w_i/\sum_{i=1}^N w_i$ we can compute ESS as 
$$\textrm{ESS}^\textrm{IS}_N = \frac{1}{\sum_{i=1}^N(\widehat{w}_i)^2} $$
If all the weights are the same (i.e. we have a sample coming from our target distribution), we have $\textrm{ESS}^\textrm{IS}_N = N$, whereas if only one of the weights is non-zero,  $\textrm{ESS}^\textrm{IS}_N = 1$. A necessary condition for this estimate to be valid however is that drawn importance samples are iid. For MCMC a similar estimate (under the same name) is available after establishing ground truth for an integral of interest. An exposition to this is given in \cite{Hoffman2014} whom we will follow. 
Given a function $\targfunc$ which we want to integrate wrt the target density $\targd$, and samples $\smp_1,\dots, \smp_N$ from a Markov Chain targeting $\targd$, that follow the distribution $\propd_{MC}$ we can use the estimate
$$\textrm{ESS}^\textrm{MC}_N = \frac{N}{1+2\sum_{i=1}^{N-1} (1-\frac{i}{N}) \acor_i^\targfunc} $$
where $\acor_i^\targfunc$ is the autocorrelation of $\targfunc$ under $\propd_{MC}$ at lag $i$.  If the autocorrelation is exactly  $0$ at each lag (i.e. all of the samples are independent), then again $\textrm{ESS}^\textrm{MC}_N=N$, as desired.  An estimate of autocorrelation is given by

$$\widehat{\acor}_i^\targfunc = \frac{1}{\Var[\targfunc(X)](N-i)} \sum_{j=i+1}^N (\targfunc(\smp_{j}) - \Expect[\targfunc(X)]) (\targfunc(\smp_{j-i}) - \Expect[\targfunc(X)])$$
In this autocorrellation estimate we make use of the ground truth values $\Expect[\targfunc(X)]$ and $\Var[\targfunc(X)]$ which can only be estimated by a long run of some Monte Carlo method when $\targd$ is a black block (for example when it is given only proportionally as a Bayesian posterior). Now as the estimate of autocorrelation is noisy for large lags, \cite{Hoffman2014} suggests to truncate the autocorrelations at the smallest cutoff $c$ with $\acor_c^\targfunc < 0.05$, yielding the estimator

$$\widehat{\textrm{ESS}}^\textrm{MC}_N = \frac{N}{1+2\sum_{i=1}^{c} (1-\frac{i}{N}) \widehat{\acor}_i^\targfunc} $$

Now as $\widehat{\textrm{ESS}}^\textrm{MC}$ is a univariate measure, \cite{Girolami2011,Hoffman2014} suggest the following approach: calculate ESS for each dimension separately and take the minimum across dimensions. Because usually in a bayesian setting one is interested in estimation of variance, the suggested procedure is to do this for an estimate of both expected value and variance of $\targd$ and take the minimum of those as the final ESS estimate.

Ironically, an ESS estimate does not exist for Sequential Monte Carlo when targeting the same density across iterations (the situation is different in a dynamic model with multiple targets). For one, it is not possible to use the estimate $\textrm{ESS}^\textrm{IS}$ because SMC induces dependency between samples. Also, $\textrm{ESS}^\textrm{MC}$ is not usable because the dependency structure in SMC, and thus the computation of autocorrelation, is much more complicated than in MCMC, as a sample at one iteration of the algorithm can spawn several samples in the next iteration.

However, using ground truth estimates $\Var[\targfunc(X)]$ and $\Expect[\targfunc(X)]$ as $\textrm{ESS}^\textrm{MC}$ does, it is possible to reproduce the worst case properties suggested by \cite{Girolami2011,Hoffman2014}. To this end, one measures maximum squared error (MaxSE) by computing, for each Monte Carlo run, squared errors with respect to $\Var[\targfunc(X)]$ and $\Expect[\targfunc(X)]$  and taking the maximum across those and across dimensions.

\subsection{Gaussian Grid}
\begin{figure}[tbp]
\begin{center}
\begin{minipage}[t]{0.5\textwidth}
\centering
\includegraphics[width=0.8\textwidth]{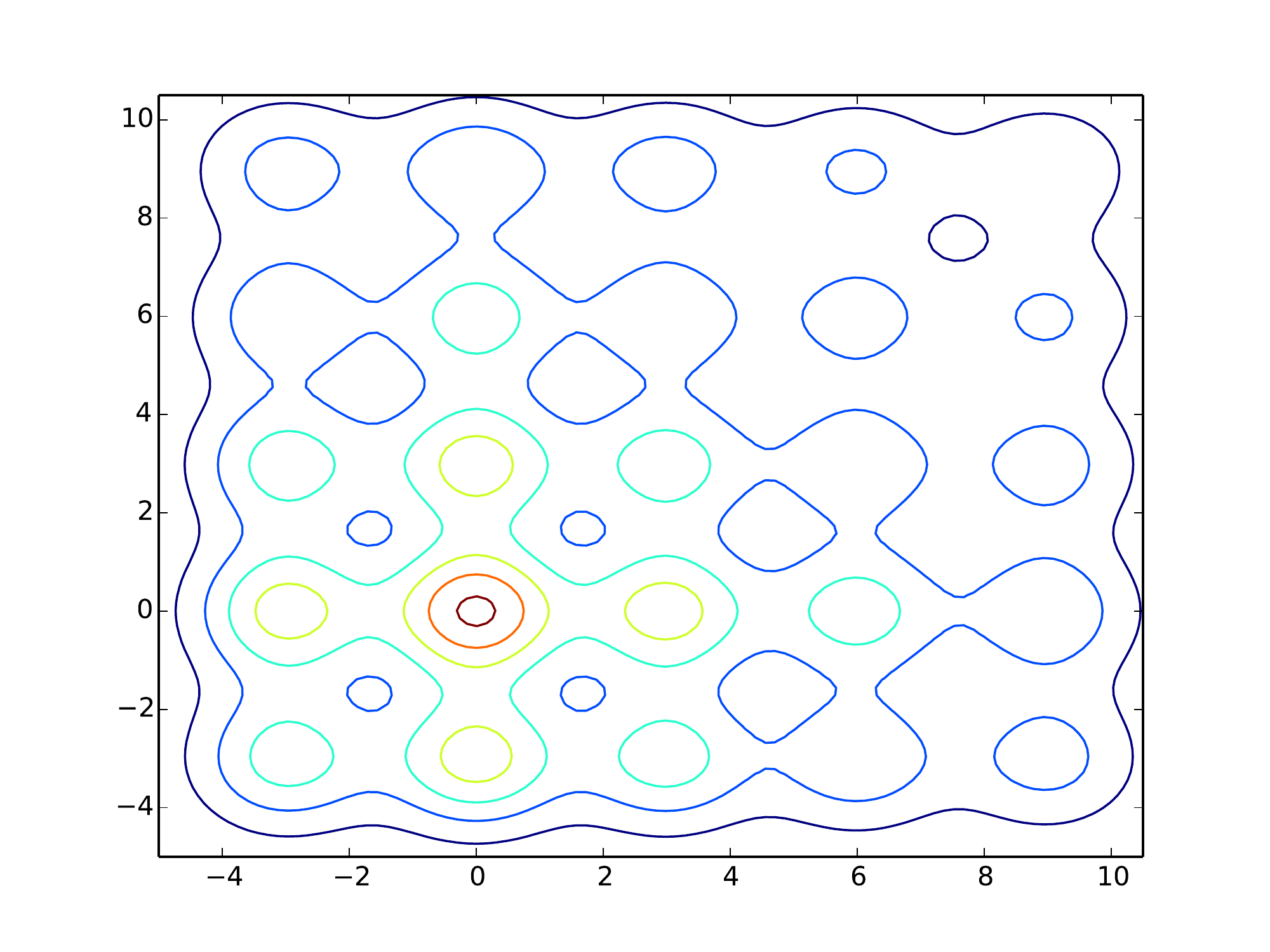} 
\caption{Gaussian mixture target: Contour plot}
\label{fig:gauss_grid_contour}
\end{minipage}\hfill
\begin{minipage}[t]{0.5\textwidth}
\centering
\includegraphics[width=0.9\textwidth]{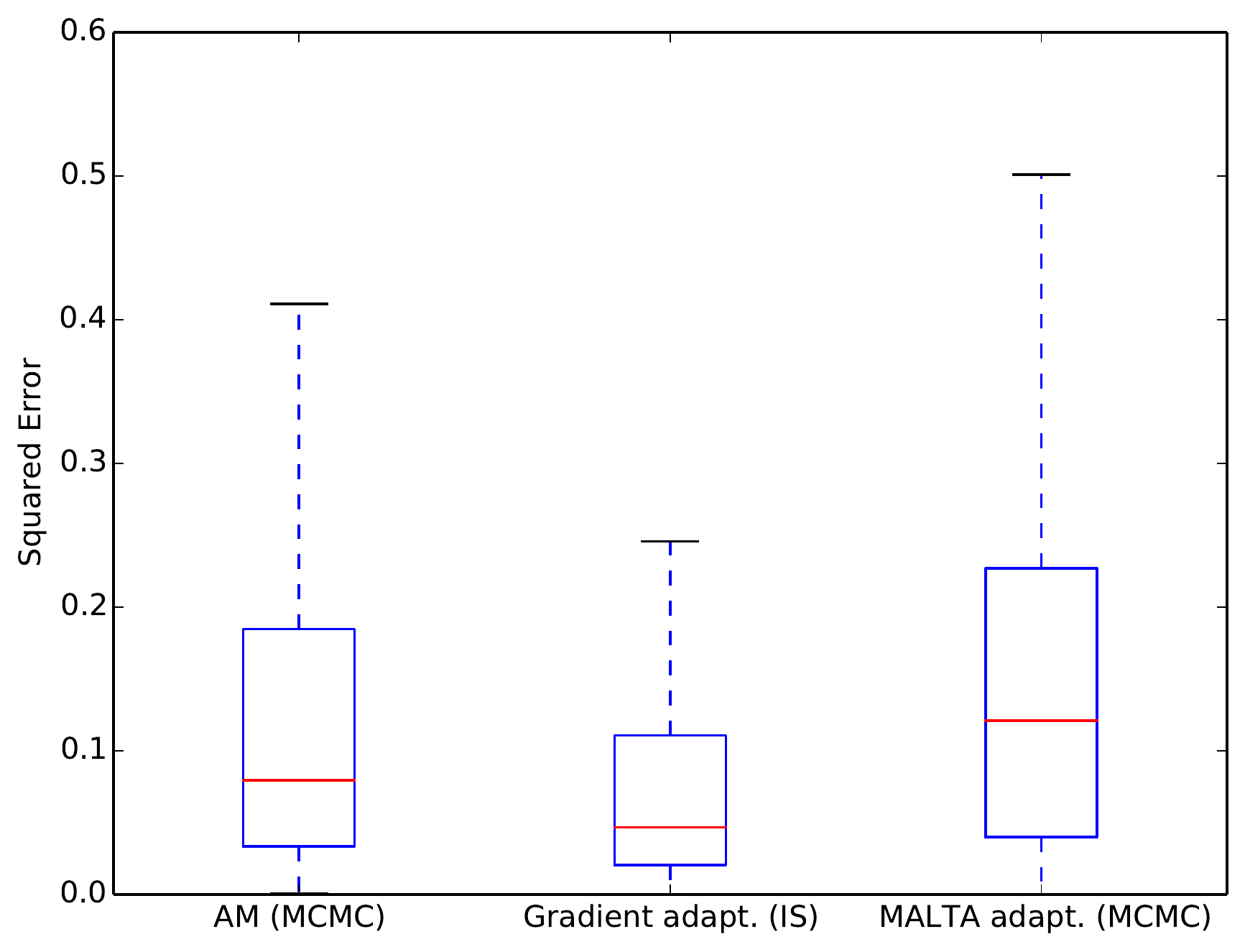} 
\caption{Gaussian Mixture target: SE performance. Algorithms not shown are widely off scale.} 

\label{fig:gauss_grid_box}
\end{minipage}

\end{center}
\end{figure}

The first target was a mixture of 25 2D Gaussians with equidistant means and mixture weights depending on distance from origin, a contour plot for which is given in Figure~\ref{fig:gauss_grid_contour}. A box plot for the squared error across different numbers of target evaluations is given in Figure~\ref{fig:gauss_grid_box}, where the box ranges from 1st to 3rd quartile and whiskers extend to $1.5*\textrm{IQR}$  (interquartile range). Hamiltonian Monte Carlo (HMC) was not plotted here, because the algorithm performed much worse than the other three candidates and proper scaling was impossible.

\begin{figure}[tbp]
\begin{center}
\includegraphics[width=0.8\textwidth]{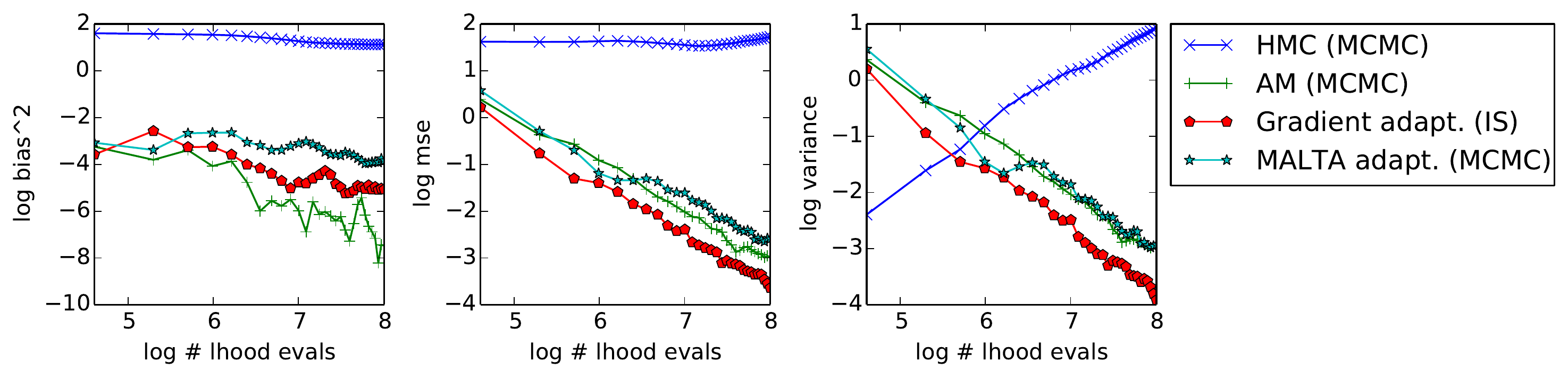}
\caption{Gaussian mixture target: Squared bias, MSE and variance as a function of number of target evaluations}
\label{fig:gauss_grid_mse}
\end{center}
\end{figure}

For this target, the performance was very close for the adaptive algorithms. Adaptive Gradient IS exhibits smallest MSE and variance (see Figure~\ref{fig:gauss_grid_mse}), but the AM algorithm is a close second. As multimodal targets are traditionally problematic for gradient informed sampling algorithms, it is interesting to see that adapting to the posterior covariance structure of the target can help mitigate problems in this case. The particularly weak performance of HMC possibly stems from the algorithm getting stuck in different modes for different runs, explaining in particular the high variance (Figure~\ref{fig:gauss_grid_mse}).

\subsection{Banana}
\begin{figure}[tbp]
\begin{center}
\begin{minipage}[t]{0.5\textwidth}
\centering
\includegraphics[width=\textwidth]{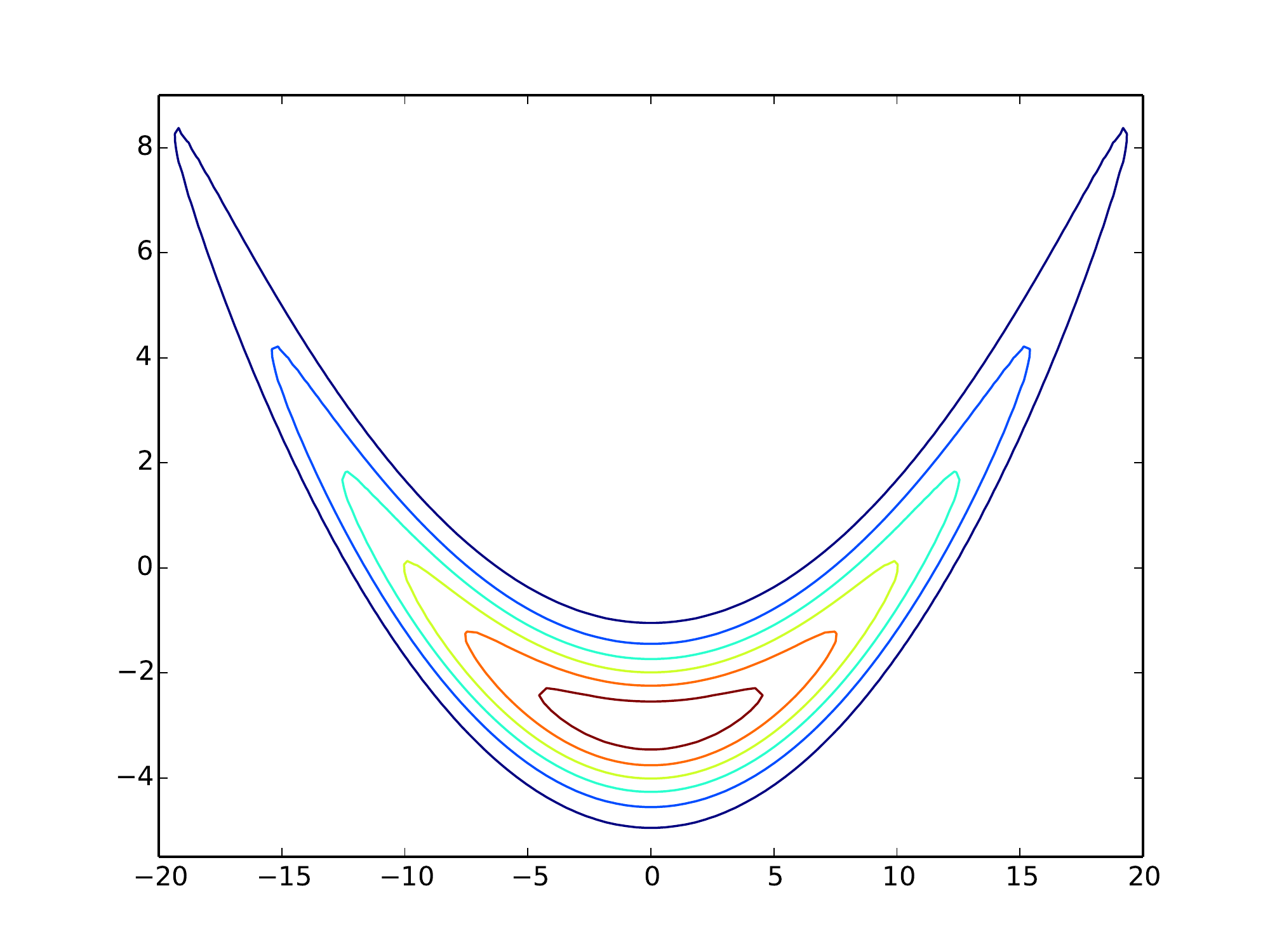} 
\caption{Banana target: Contour plot}
\label{fig:banana_contour}
\end{minipage}\hfill
\begin{minipage}[t]{0.5\textwidth}
\centering
\includegraphics[width=\textwidth]{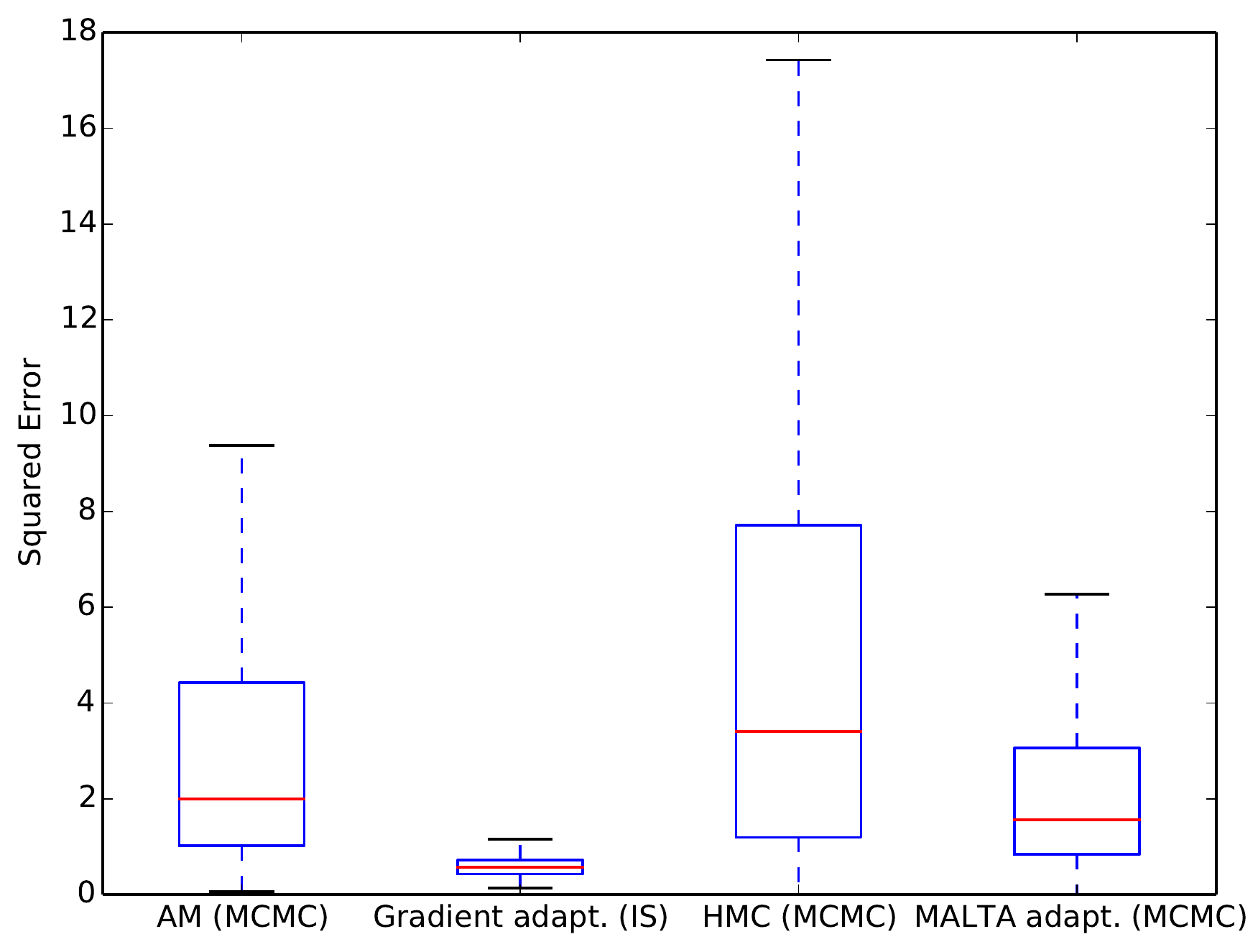} 
\caption{Banana target: SE performance} 

\label{fig:banana_box}
\end{minipage}

\end{center}
\end{figure}

\begin{figure}[tbp]
\begin{center}
\includegraphics[width=0.8\textwidth]{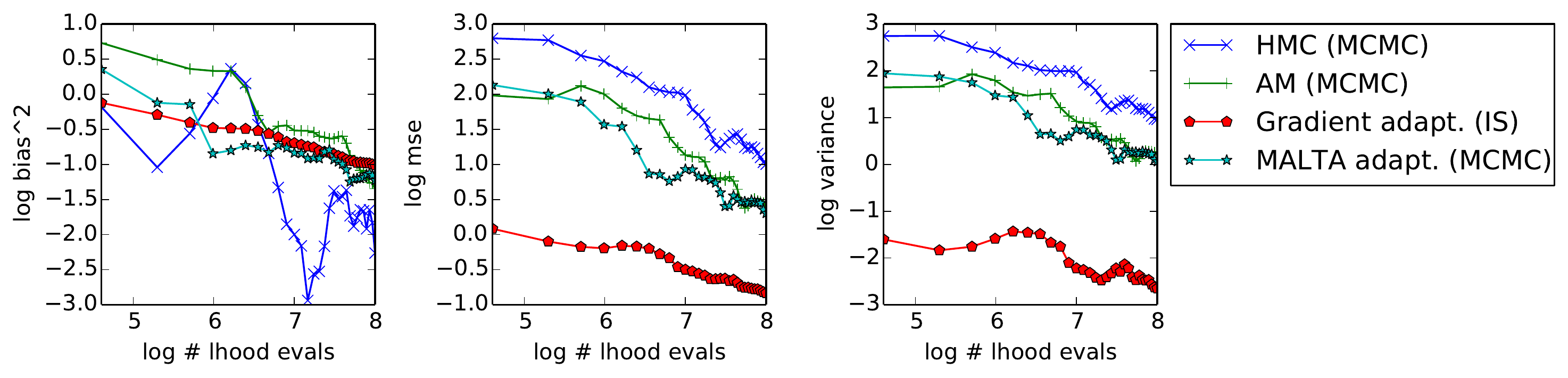}
\caption{Banana target: Squared bias, MSE and variance as a function of number of target evaluations}
\label{fig:Banana_mse}
\end{center}
\end{figure}

The 2D Banana shaped unnormalized measure was given by $f(x) = \textrm{exp}(-\frac{1}{2s}x_1^2-\frac{1}{2}(x_2-b(x_1^2-s))^2$. The measure is determined by parameters $b$ and $v$ which where set to $100$ and $0.03$, respectively (see Figure~\ref{fig:banana_contour} for a contour plot). For unimodal targets, gradient based samplers are traditionally strong, though that advantage might be irrelevant when we adapt a scale matrix for a simple random walk algorithm. As is evident from Figures~\ref{fig:banana_box} and \ref{fig:Banana_mse}, the simple AM algorithm actually is competitive with adaptive T-MALA for this target. However, adaptive Gradient IS again shows gains here when compared to the other samplers. This is particularly encouraging since  T-MALA and Gradient IS are using the exact same drift function and T-MALA adapts more parameters than Gradient IS does. If the remaining parameters of Gradient IS can be adapted automatically, further improvements might be possible.

\subsection{Mixture of T distributions}
\begin{figure}[tbp]
\begin{center}
\begin{minipage}[t]{0.5\textwidth}
\centering
\includegraphics[width=\textwidth]{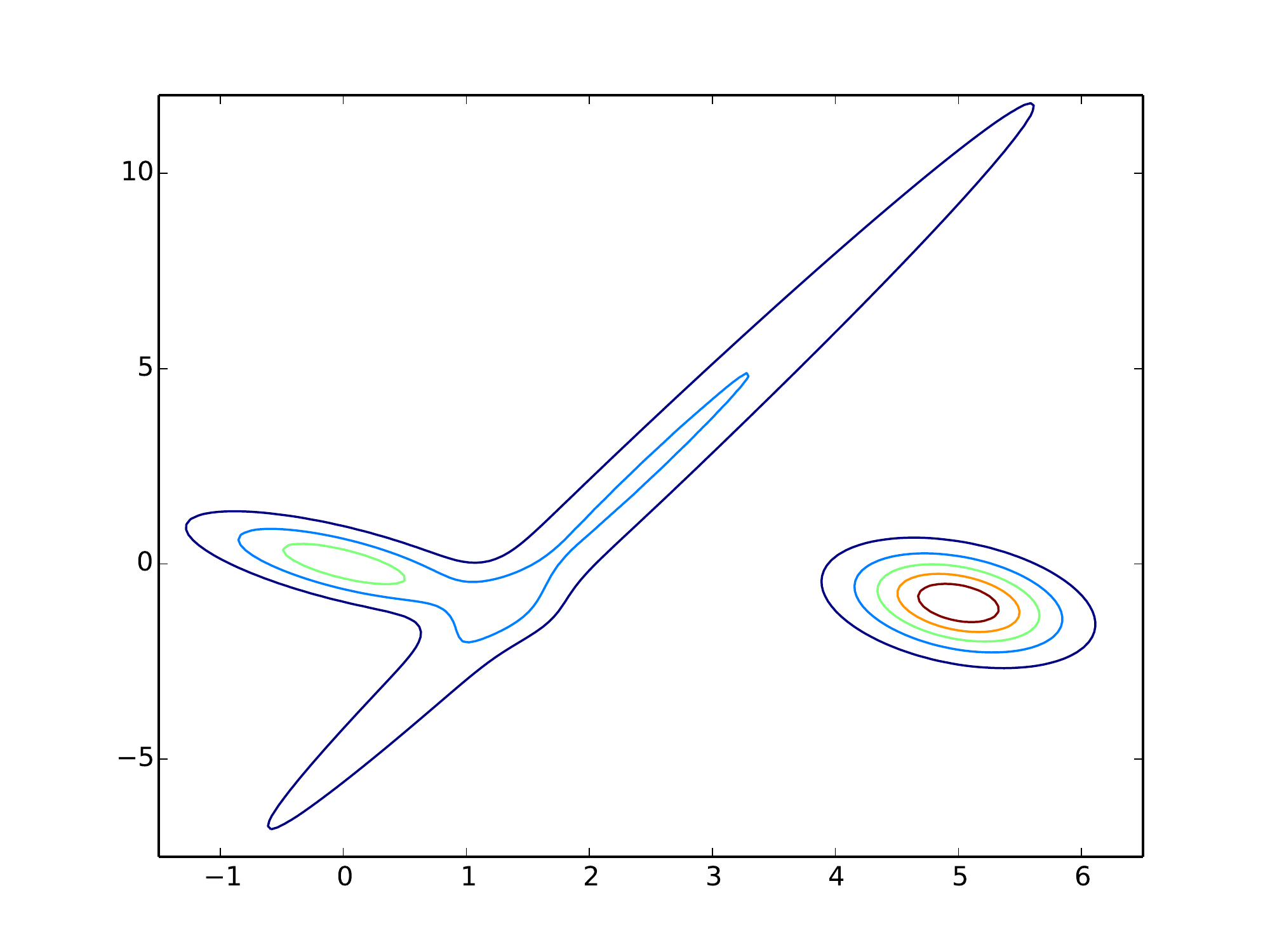} 
\caption{10 D $t$-Mixture target: Contour plot of the marginal density of the last two dimensions}
\label{fig:t_mixt_contour}
\end{minipage}\hfill
\begin{minipage}[t]{0.5\textwidth}
\centering
\includegraphics[width=0.8\textwidth]{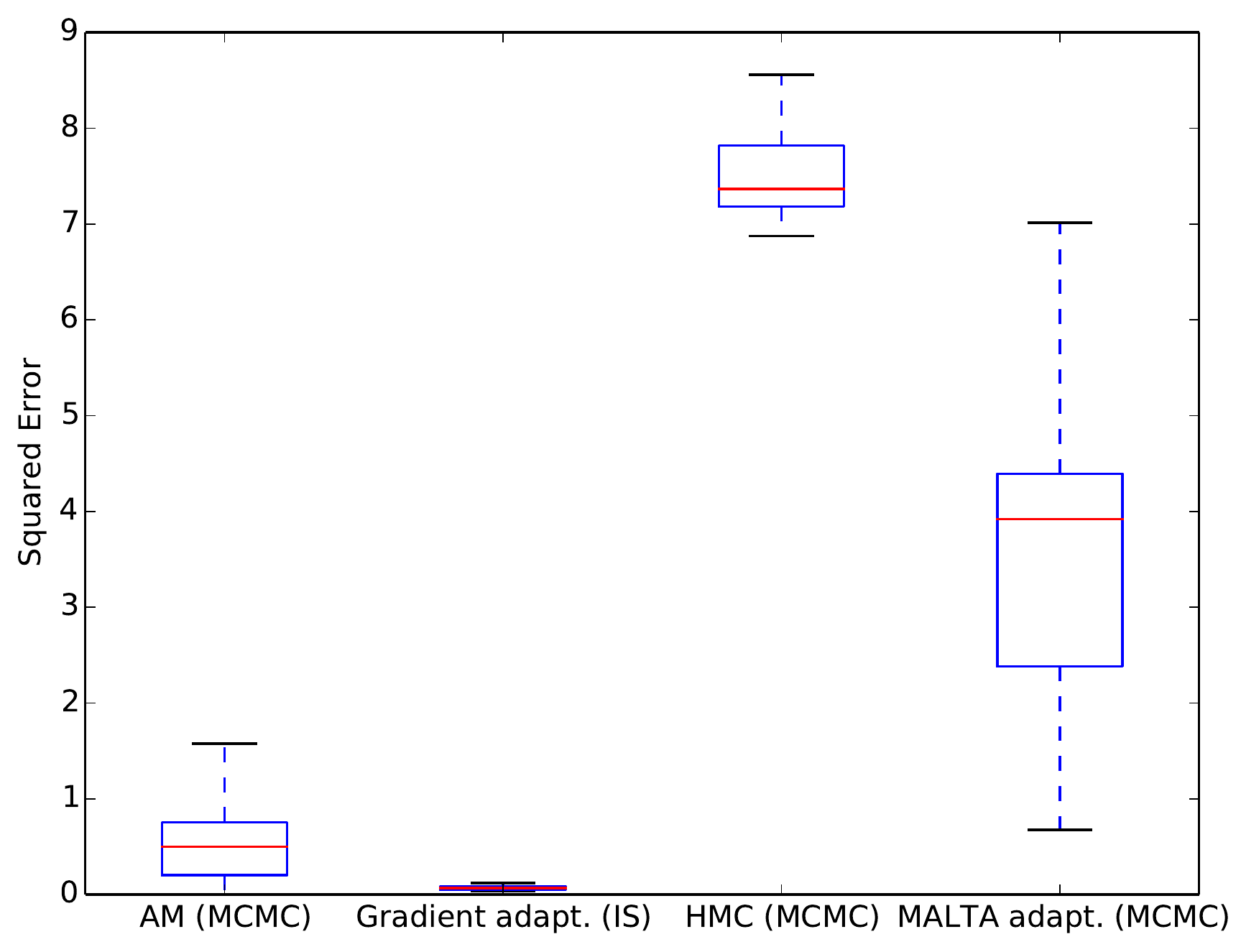} 
\caption{$t$-Mixture target: SE of mean estimates, averaged across dimensions} 

\label{fig:t_mixt_box}
\end{minipage}

\end{center}
\end{figure}

\begin{figure}[tbp]
\begin{center}
\includegraphics[width=0.5\textwidth]{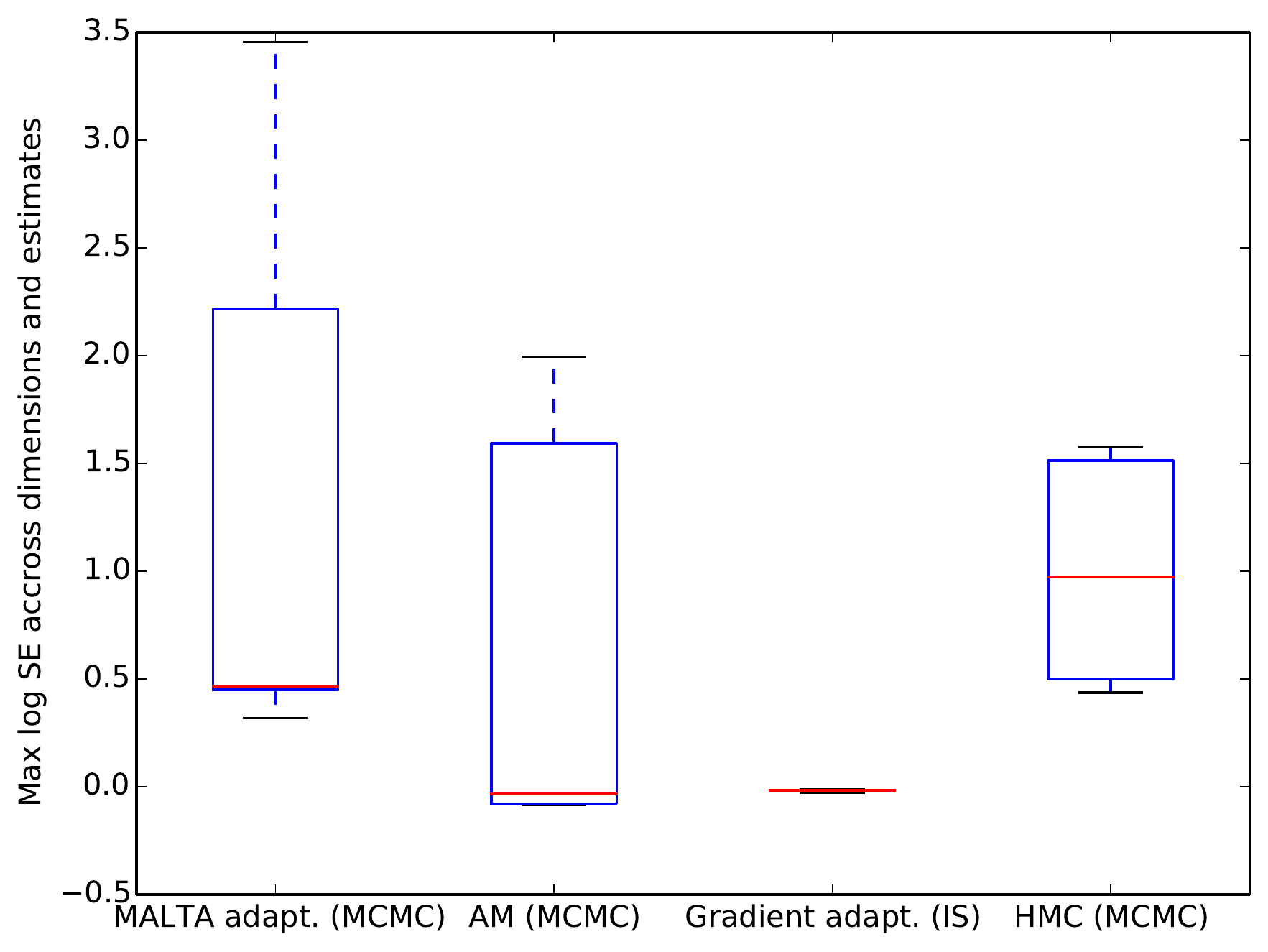}
\caption{$t$-Mixture target: Squared bias, MSE and variance as a function of number of target evaluations}
\label{fig:t_mixt_mse}
\end{center}
\end{figure}

The third target considered was a mixture of three $10$D multivariate $t$-distributions with different means, $10$ degrees of freedom and varying mixture weights. The scale matrices where drawn from an Inverse Wishart distribution with identity scale matrix and $10$ degrees of freedom, yielding mixture components with strong correlations among dimensions.  A contour of the marginal density of the last two coordinates is given in Figure~\ref{fig:t_mixt_contour}. This case was considered because $t$-distributions have heavier tails than Gaussians. As a standard rule of thumb, IS proposal distributions should have heavier tails than the target to ensure finite variance of the estimate \cite{Robert2004}. With a mixture of $t$-distributions, I wanted to experimentally check for any problems stemming from the Gaussian proposal used in all algorithms. GRIS is better when averaging squared error across dimensions (Figure \ref{fig:t_mixt_box}). Also, when when comparing maximum log squared errors GRIS is clearly an improvement over previous algorithms (Figure \ref{fig:t_mixt_mse}). 

\subsection{German Credit Dataset: Logistic regression}
The German Credit dataset was used with the Logistic Regression model developed in \cite{Hoffman2014}. This model exhibited a $25$D posterior distribution which allowed for a Laplace approximation. To find ground truth, I collected ordinary Importance Samples from a mixture between the Laplace Approximation and the sample approximation with slightly increased scale, a method known as defensive Importance Sampling \cite{Owen2000}. The Effective Sample Size of this approximation was over $66,000$.

\begin{figure}[tbp]
\begin{center}
\begin{minipage}[t]{0.5\textwidth}
\centering
\includegraphics[width=0.8\textwidth]{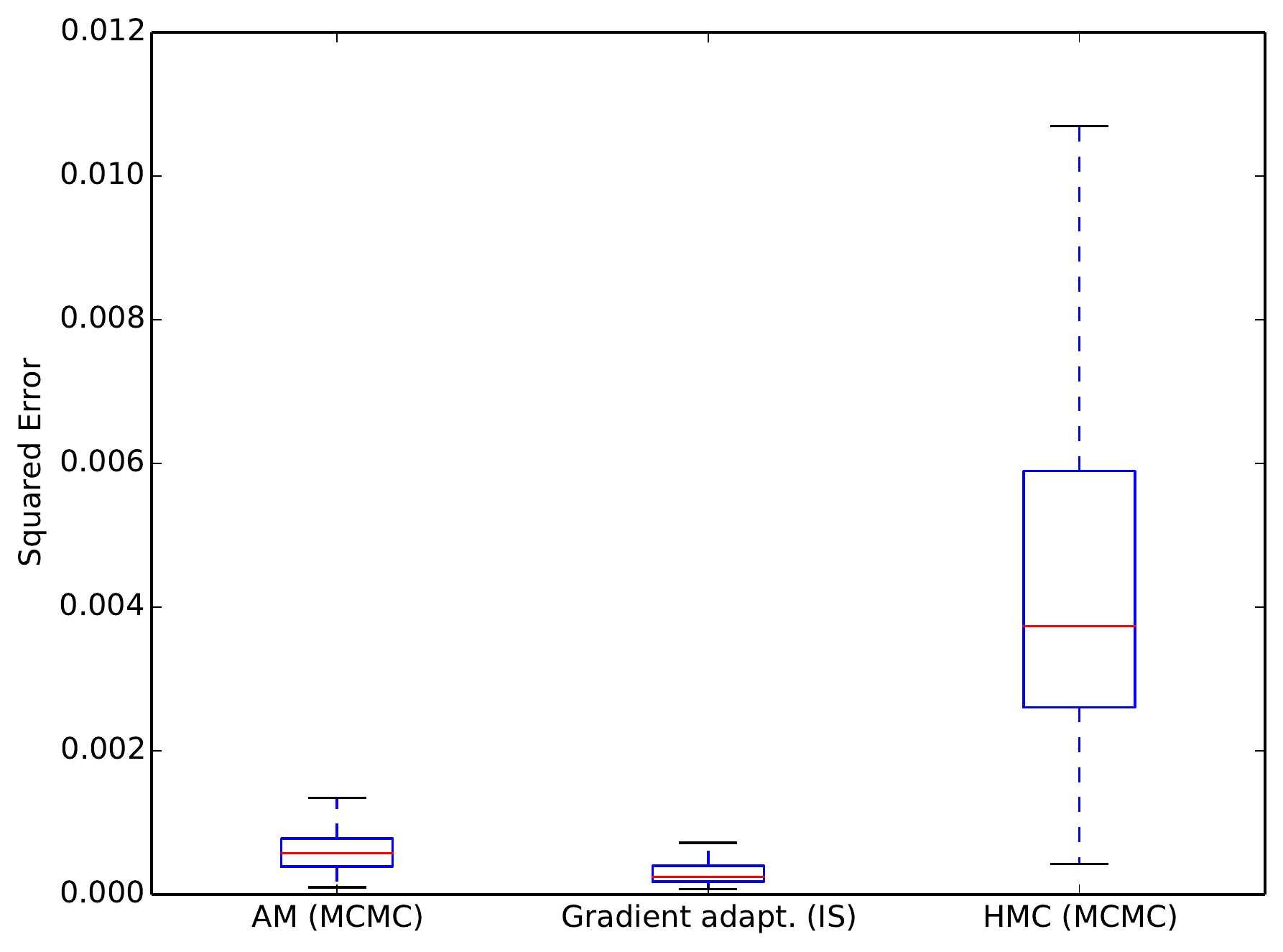} 
\caption{Logistic regression: Squared errors of mean estimate averaged across dimensions. Algorithms not shown are widely off scale.}
\label{fig:cred_worst}
\end{minipage}\hfill
\begin{minipage}[t]{0.5\textwidth}
\centering
\includegraphics[width=0.75\textwidth]{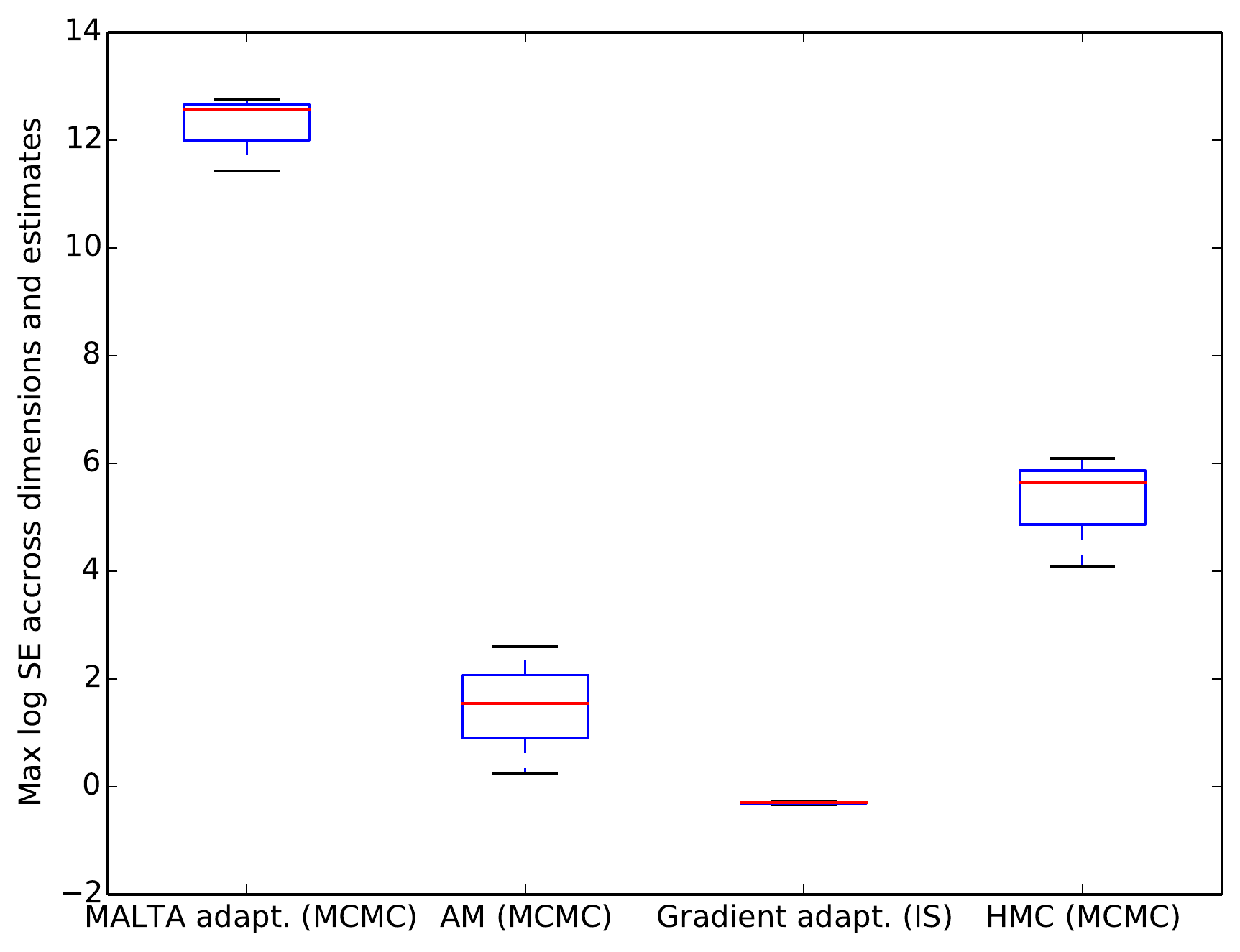} 
\caption{Logistic regression: Maximum log SE across estimates of posterior variance and mean and across dimensions.} 

\label{fig:cred_worst}
\end{minipage}

\end{center}
\end{figure}

\subsection{Evidence Estimation}
\begin{figure}[tbp]
\begin{center}
\centering
\includegraphics[width=0.8\textwidth]{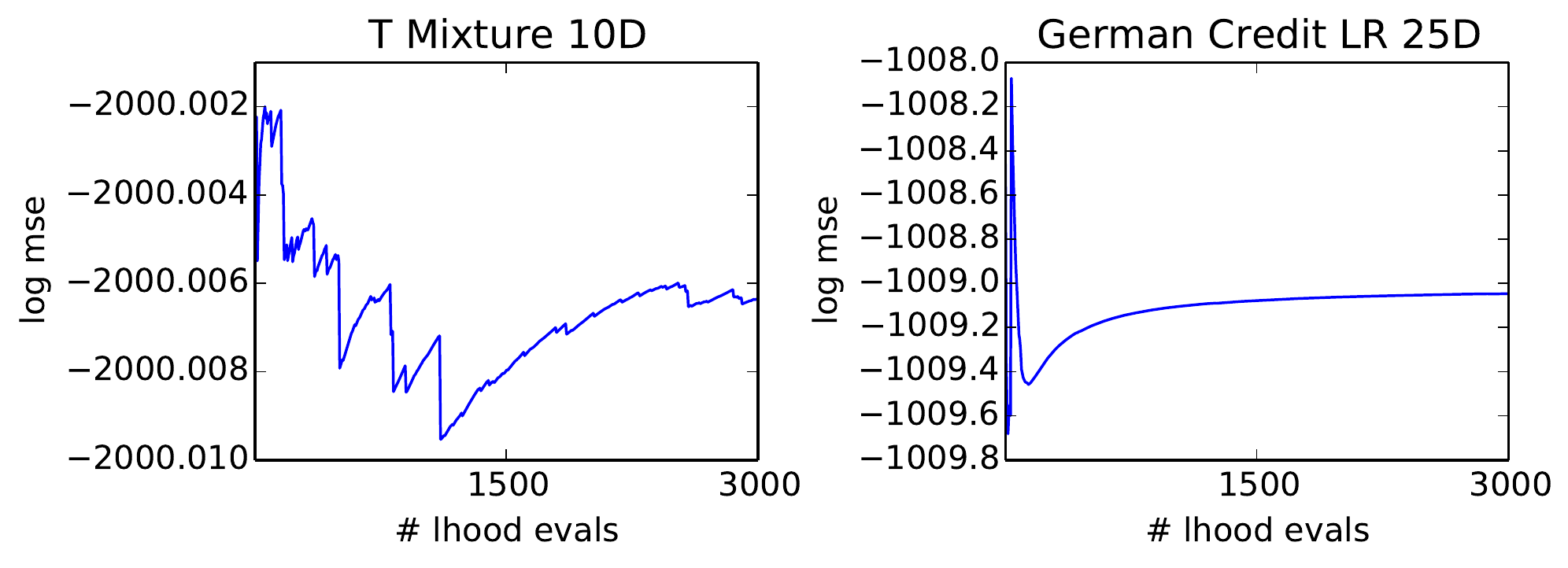} 
\caption{Evidence estimates}
\label{fig:ev}

\end{center}
\end{figure}

Evidence estimates using GRIS quickly stabilized. I assesed MSE of evidence estimates for the $t$-Mixture target and the Logistic Regression model  (Figure \ref{fig:ev}). The log evidences where $-1000$ and $-504$, respectively.

\section{Conclusions}
In this paper, I presented Gradient IS, a variant of Sequential Monte Carlo. The algorithm uses gradient information and  a covariance matrix adapted to collected posterior samples to construct a multivariate normal proposal distribution for SMC with static target distributions. GRIS was shown to give very good performance in posterior sampling and provide stable estimates for the normalizing constant of the target distribution (also known as model evidence).

\subsubsection*{Acknowledgments}
I am grateful to Christian Robert, Marco Banterle and Nicolas Chopin for helpful discussions. Also, Patrick J\"ahnichen has proofread the manuscript and hinted at some of the errors and possible improvements in presentation.


\bibliography{library}
\bibliographystyle{unsrt}

\end{document}